\documentclass[runningheads]{llncs}
\usepackage{marvosym}
\usepackage{authblk}
\usepackage{xcolor}

\usepackage{graphicx}
\usepackage{epstopdf}
\usepackage{booktabs}
\usepackage{diagbox} 
\usepackage{multirow}
\usepackage{amstext}
\usepackage{hyperref}
\usepackage{amsfonts, mathtools}
\usepackage{graphicx} 
\usepackage{caption} 
\usepackage{subcaption}
\begin{document}
\title{Kolmogorov-Arnold Networks (KAN) for Time Series Classification and Robust Analysis}
\titlerunning{KAN for Time Series Classification and Robust Analysis}

\author{
Chang Dong \inst{1}
\and
Liangwei Zheng  \inst{1}
\and 
Weitong Chen \thanks{Corresponding Author} \inst{1}
}
\authorrunning{Chang et al.}
\institute{\textsuperscript{1}The University of Adelaide\\
\email{\{chang.dong, liangwei.zheng, weitong.chen\}@adelaide.edu.au}\\
}

\maketitle              
\begin{abstract}

Kolmogorov-Arnold Networks (KAN) has recently attracted significant attention as a promising alternative to traditional Multi-Layer Perceptrons (MLP). Despite their theoretical appeal, KAN require validation on large-scale benchmark datasets. 
Time series data, which has become increasingly prevalent in recent years, especially univariate time series are naturally suited for validating KAN. Therefore, we conducted a fair comparison among KAN, MLP, and mixed structures. The results indicate that KAN can achieve performance comparable to, or even slightly better than, MLP across 128 time series datasets. 
We also performed an ablation study on KAN, revealing that the output is primarily determined by the base component instead of b-spline function. Furthermore, we assessed the robustness of these models and found that KAN and the hybrid structure MLP\_KAN exhibit significant robustness advantages, attributed to their lower Lipschitz constants.  This suggests that KAN and KAN layers hold strong potential to be robust models or to improve the adversarial robustness of other models.

\keywords{Kolmogorov-Arnold Networks\and Time-Series \and Adversarial Attack }

\end{abstract}
%
%

\section{Introduction}

In recent years, time-series analysis has become increasingly prevalent in various fields, including healthcare\cite{forestier2018surgical}, human activity recognition\cite{nweke2018deep}, remote sensing\cite{pelletier2019temporal} etc. Among these, time series classification (TSC) is one of the key challenges in time series analysis. 
Due to rapid advancements in machine learning research, many TSC algorithms have been proposed.  Before the prevalence of deep learning, numerous effective algorithms existed. For instance, NN-DTW \cite{bagnall2017great} is based on measuring the general similarity of whole sequences using different distance metrics. Other methods included identifying repeated shapelets in sequences as features \cite{hills2014classification}, discriminating time series by the frequency of repetition of some subseries \cite{schafer2015boss}, and using ensemble methods \cite{7837946}. However, these approaches often faced limitations, as they were either difficult to generalize to various scenarios or had high time complexity.

In recent years, deep learning-based methods have achieved notable success in various fields such as image recognition and natural language processing. Consequently, many techniques from these domains have been adapted for time series classification, including Convolutional Neural Networks (CNNs)\cite{ismail2020inceptiontime}, Recurrent Neural Networks (RNNs)\cite{karim2017lstm}, and Transformers\cite{wen2022transformers}. Even large model-based classification has found extensive application in time series classification\cite{zhou2023one}.


The development of neural networks is fundamentally rooted in the concept of the multi-layer perceptron (MLP). Regardless of their complexity, neural networks retain an architecture similar to that of MLP. According to the Universal Approximation Theorem (UAT), any function can be approximated by a finite number of single-layer perceptrons\cite{hornik1989multilayer}. This theorem underpins the capability of MLP to model and fit complex distributions.
Recently, Liu, et al have proposed a new paradigm for neural networks called Kolmogorov-Arnold Networks (KAN)\cite{liu2024kan}, which contrasts with traditional MLP-based neural networks. Unlike MLP, KAN is based on Kolmogorov-Arnold Theory (KAT)\cite{kolmogorov1961representation} and explicitly defines the model size required for fitting. Both MLP and KAN have analogous structures: in MLP, neuron outputs undergo a linear transformation followed by activation before being passed to the next neuron, whereas in KAN, the edges serve as learnable activation functions, followed by a linear transformation before passing to the next neuron. This learnable activation function makes KAN a potential competitor to MLP neural networks. 
However, KAN has only been validated within formulas constructed in the physical domain and has not been tested on large-scale datasets, leaving its scalability unproven. Univariate time series are inherently well-suited to KAN' inputs, making them excellent candidates for validation. Furthermore, the robustness of TSC  has garnered significant attention in recent years\cite{dong2024boosting,dong2023swap,fawaz2019adversarial,karim2020adversarial}. As a new architecture, KAN' robustness has not yet been studied. Given these circumstances, to validate the performance and robustness of KAN in TSC tasks, we conducted the following work:
\begin{itemize}
    \item We performed a fair comparison across 128 UCR datasets among KAN, MLP, KAN\_MLP (KAN with the last layer replaced by MLP), MLP\_KAN (MLP with the last layer replaced by KAN) under identical configurations, and MLP\_L (MLP with the same number of parameters). We found that KAN could achieve performance comparable to MLP.\\
    
    \item We conducted an ablation study to investigate the roles of the base and B-spline functions. The results indicated that the output values were predominantly determined by the base function. Additionally, we observed that in the absence of the base function, spline functions with large grid sizes were difficult to optimize.\\
    
    \item We assessed the robustness of KAN by comparing it with other models. Our findings revealed that KAN exhibited superior adversarial robustness due to its lower Lipschitz constant.\\
        
    \item We observed an anomalous phenomenon that KAN with higher grid sizes demonstrated greater robustness despite having a higher Lipschitz constant. We provided a reasonable hypothesis for this observation in the final section.

\end{itemize}

\section{Background}


\subsection{Kolmogorov-Arnold representation}

KAN is inspired by Kolmogorov-Arnold representation theory (KAT). It states that any multivariate continuous function defined in a bounded domain can be represented as a finite composition of continuous functions of a single variable and the binary operation of addition. Specifically, if $f$ is a continuous function on a bounded domain $D \subset \mathbb{R}^n$, then there exist continuous functions $\phi_{ij}$ and $\psi_i$ such that:

\begin{equation}
f(x_1, x_2, \ldots, x_n) = \sum_{i=1}^{2n+1} \psi_i \left( \sum_{j=1}^n \phi_{ij}(x_j) \right), 
\end{equation}
where $\phi_{ij}: [0, 1] \rightarrow \mathbb{R}$ and $\psi_i: \mathbb{R} \rightarrow \mathbb{R}$. It transforms the task of learning a multivariable function into learning a finite number of univariable functions. Compared to MLP, it explicitly provides the number of one-dimensional functions needed for fitting. However, these univariable functions could be non-smooth or even fractal, making it theoretically feasible but practically useless. Nevertheless, Liu et al.\cite{liu2024kan} found that, by analogy to MLP in neural networks, KAN need not be limited to two layers and finite width to fit all non-linearities. Furthermore, most natural functions tend to be smooth and have sparse structures. These insights suggest that an scalable KAN could become a strong competitor to MLP.

\subsection{Adversarial Attack}
Adversarial attacks involve applying carefully crafted small perturbations $r \in \mathbb{R}^d$ to input data $x \in \mathbb{R}^d$, leading to significant changes in a model’s output, such as fooling a classifier $f: x \rightarrow \mathbb{R}^m$ with the goal of altering the predicted label.
\begin{equation}
\begin{split}
    \text{argmax}\ \{f(x)\} \ne \text{argmax}\ \{f(x_{adv})\}, \\
    x_{adv} = x + r,\ \text{s.t.} \ ||r||^2 \ll ||x||^2.
\end{split}
\end{equation}
here, the perturbation \( r \) is small in magnitude relative to \( x \) as indicated by their norms. We normally apply these perturbations to test whether the model can be fooled, thus assessing its robustness. To consider the worst-case scenario, we typically implement the gradient attacks which require knowledge of all the information about the model and the data. Among them, the most widely used method is the Projection Gradient Descent (PGD)\cite{madry2017towards}, the gradient-based iterative attack method, which is the most effective method to evaluate the robustness of models against gradient attacks.
This can be characterized by:
\begin{equation}
x_{adv}^{(t+1)} = Clip_{x, \epsilon} \{ x_{adv}^{(t)} + \alpha \cdot \text{sign}(\nabla_{x_{adv}^{(t)}} \mathcal{L}(x_{adv}^{(t)}, y)) \},
\end{equation}
here, \( t \) is the iteration index, \( Clip_{x, \epsilon} \{ \cdot \} \) ensures that \( x_{adv}^{(t+1)} \) remains within \( \epsilon \) of the original input \( x \). This method iteratively adjusts \( x_{adv} \) to maximize the loss function in the direction that moves it away from its original prediction $y$, while ensuring \( x_{adv} \) stays within a small perturbation distance \( \epsilon \) from \( x \).

\subsection{Local Lipschitz Constant}
A function \( f : \mathbb{R}^m \to \mathbb{R}^n \) is defined to be \( \ell_f \)-locally Lipschitz continuous at radius \( r \) if for each \( i = 1, \ldots, n \), and \( \forall \ \|x_1-x_2\| \leq r \), the following holds:
\begin{equation}
\| f(x_1) - f(x_2) \| \leq \ell_f \| x_1 - x_2 \|
\end{equation}
where \( \ell_f \) is the local Lipschitz constant. Hereafter, we will refer to it simply as the Lipschitz constant. The Lipschitz constant is directly linked to perturbation stability, which in turn relates to robustness\cite{tholeti2022robust}.

\section{Methodology}

\subsection{Kolmogorov-Arnold Networks (KAN)}
\begin{figure}[h]
    \centering
    \includegraphics[width=1\textwidth]{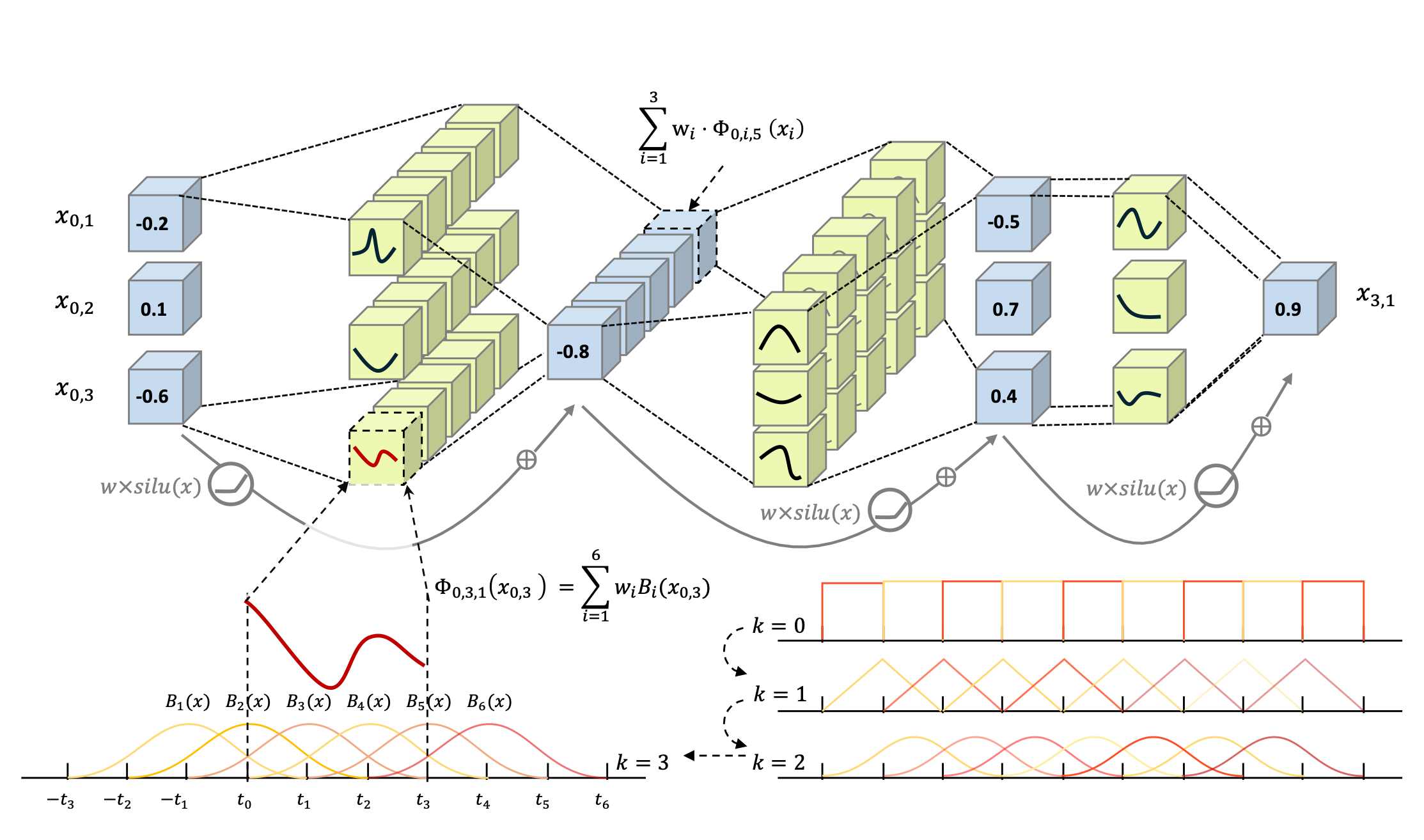} 
    \caption{A three-layer KAN structure with the architecture [3-5-3-1].} 
    \label{fig:mainpic} 
\end{figure}
Assume there is a data distribution \( D \subseteq \mathbb{R}^d \times \mathbb{R}^m \). Our objective is to learn a function \( f: x \in \mathbb{R}^d \rightarrow y \in \mathbb{R}^m \) such that the following risk is minimized as $\hat{R}(f) = \frac{1}{n} \sum_{i=1}^{n} \| y_i - f(x_i) \|$. The purpose of the KAN is to learn such a representation of \( f \), thereby minimizing the objective loss. The original KAN used a two-layer structure, while Liu, et al. \cite{liu2024kan} extended to arbitrary width and depth. In contrast to MLP, the activation function are placed on edges instead of the neurons, KAN use $3^{rd}$-order B-spline ($k = 3$) functions for fitting, which allows learning sophisticated activation function by controlling the weight of each basis. In this case, the neuron $q$ in the layer $l+1$ can be represented as :
\begin{equation}
x_{l+1, q}^{spline} = \sum_{p=1}^{n} w_{p,q}^{spline}\cdot \Phi_{l,q,p}(x_{l,p}) = \sum_{p=1}^{n} w_{p,q}^{spline}\cdot \sum_{i=1}^{k+G} w_i \cdot B_i(x_{l,p}),
\end{equation}
where $x_{l,p}$ is the input from an arbitrary neuron $p$ in the previous layer $l$. The input from all $n$ neurons in the previous layer $l$ undergoes a nonlinear transformation produced by a learnable B-spline combination, where $G$ is the grid size which determines the number of B-spline bases ($k+G$). This is followed by a weighted summation to obtain the $q^{th}$ output of $x_{l+1, q}^{spline}$. Additionally, KAN introduce a base function similar to residual connections, using weighted $silu$, to stabilize optimization, which can be represented as:
\begin{equation}
x_{l+1, q}^{base} = \sum_{p=1}^{n} w_{p,q}^{base}\cdot silu(x_{l,p}) = \sum_{p=1}^{n} w_{p,q}^{base}\cdot \frac{x_{l,p}}{1+e^{-x_{l,p}}}
\end{equation}
Therefore, the output of the $q^{th}$ neuron in layer $l+1$ can be represented as:
\begin{equation}
x_{l+1, q}  = x_{l+1, q}^{spline} + x_{l+1, q}^{base}
\end{equation}
For a multi-layer KAN, the final output can be represented as a nested structure of layers:
\begin{equation}
f(\mathbf{x} ) = f(x_1, x_2, ..., x_n) = \Psi_L \circ \Psi_{L-1} ...\circ \Psi_1 \circ \mathbf{x} 
\end{equation}
where $\Psi_l$ denotes the $l^{th}$ layer, which includes the combination of the above two operations: a $spline$ transformation and a base activation $silu$. Fig.~\ref{fig:mainpic} illustrates a three-layer KAN structure with the architecture [3-5-3-1], clearly depicting how KAN operate.

\subsection{KAN for time series classification}
We constructed classifiers using KAN, similar to the structure shown in fig.~\ref{fig:mainpic}. Due to the setting of the B-spline fitting interval being $[-1, 1]$, the data distribution outside this interval will not achieve an effective fitting. Instead of directly adopting the method proposed by Liu et al., which suggested updating the grid interval according to data distribution, We employed a more straightforward approach. We fixed the B-Spline grid interval to $[-1, 1]$ throughout the process, and applied batch normalization to keep the distribution within $[-1, 1]$ in each KAN Layer, to ensure the data distribution conforms to the grid and optimize the training process. Thus, to build a KAN for TSC, we adopted a 3-layer structure with the output transformed to the number of classes, having an architecture of [d-d-128-m], where $d$ is the sequence length and $m$ is the number of classes. Meanwhile, We compared KAN with MLP that had the same number of parameters and neurons per layer, as well as networks where the last layer of KAN was replaced with MLP (KAN\_MLP) and the last layer of MLP was replaced with KAN (MLP\_KAN). The experimental design is shown in tab.~\ref{table:Model_configurations}.

\begin{table}[h!]
\centering
\caption{Model Architectures and Parameters(G=5, k= 3 for all B-splines)}
\label{table:Model_configurations}
\begin{tabular}{c|c|c|c}
\toprule
\textbf{Networks} & \textbf{Architecture} & \textbf{Activation} & \textbf{Parameters} ($\approx$) \\
\midrule
KAN & [d-d-128-m] & Silu, B-Spline & $(2 + G + k)\cdot d^2 + (258 + 128 \cdot (G+k)) \cdot d $ \\
MLP\_I & [d-d-128-m] & Relu & $d^2 + 131d$ \\
MLP\_II & [d-10d-128-m] & Relu & $10d^2 + 1310d$ \\
KAN\_MLP & [d-d-128-m] & Relu, Silu, B-Spline & $(2 + G + k) \cdot d^2 + 130d$ \\
MLP\_KAN & [d-d-128-m] & Relu, Silu, B-Spline& $d^2 + (2 + G + k)\cdot128 d $ \\
\bottomrule
\end{tabular}
\end{table}

\section{Experiment Settings}

 \textbf{Dataset:}
 We applied the UCR2018 datasets \cite{dau2019ucr} to evaluate these models. The UCR Time Series Archive encompasses 128 datasets, which are all univariate. These datasets span a diverse range of real-world domains, including healthcare, human activity recognition, remote sensing and more. Each dataset comprises a distinct number of samples, all of which have been pre-partitioned into training and testing sets. Reflecting the intricacies of real-world data, the archive includes datasets with missing values, imbalances, and those with limited training samples. \\
 
\noindent{\textbf{Evaluation Metrics:}
We used the accuracy and the F1 score to assess the performance of all models in tab.~\ref{table:Model_configurations}. During adversarial attacks, we evaluate the robustness of the models using the Attack Success Rate (ASR).\\}

\noindent{\textbf{Experiment setup:}
Our experiments were executed on a server equipped with Nvidia RTX 4090 GPUs, 64 GB RAM, and an AMD EPYC 7320 processor.\\}

\noindent{\textbf{Parameter setting:}
In our experiments, we utilized the open-source GitHub project efficient-KAN\footnote{\href{https://github.com/Chang-George-Dong/KAN-for-TSC.git}{Our Github}, and \href{https://github.com/Blealtan/efficient-kan}{ efficient-KAN}. } to replace the original CPU-based KAN architecture proposed by Liu et al. \cite{liu2024kan}.  This modification, along with switching the optimizer to AdamW from BFGS, allowed for faster training speeds. We set the dropout rate to 0.1 and trained all the models for 1000 epochs. The learning rate was initialized at 1e-2 and decayed to $90\%$ of its previous value every 25 epochs. Especially for KAN, we used a weight decay of 1e-2, set L1 regularization for the weights to 0, and entropy regularization to 1e-5.
For adversarial attacks, we employed the PGD with non-targeted attacks. The perturbation magnitude eps($\epsilon$) is set at [0.05, 0.1, 0.25, 0.5], with a step size of 0.01 times eps and 100 iterations for each attack.}

\section{Result}

\subsection{Performance Comparison}
\begin{figure}[h]
    \centering
    \begin{subfigure}[b]{0.45\textwidth}
        \centering
        \includegraphics[width=\textwidth]{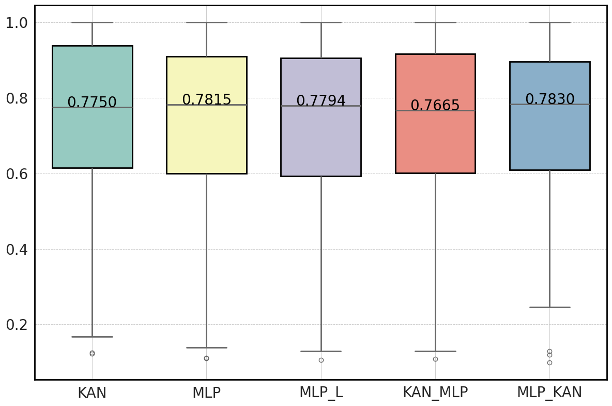} 
        \caption{Test Accuracy of five models} 
        \label{fig:accuracy} 
    \end{subfigure}
    \hfill
    \begin{subfigure}[b]{0.45\textwidth}
        \centering
        \includegraphics[width=\textwidth]{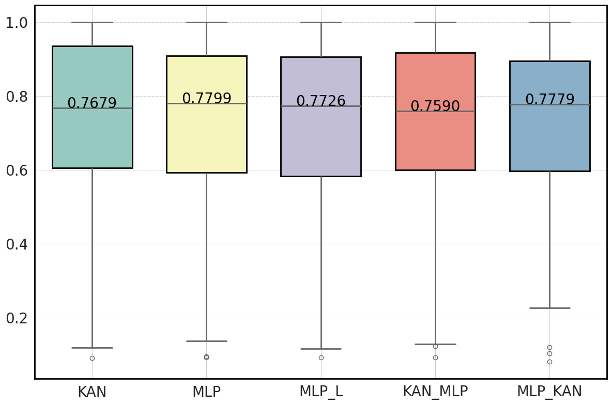} 
        \caption{Test F1 Score of five models} 
        \label{fig:f1} 
    \end{subfigure}
    \caption{Performance comparison of five models across 128 datasets} 
    \label{fig:Main_Result} 
\end{figure}

\begin{figure}[h]
    \centering
    \includegraphics[width=1\textwidth]{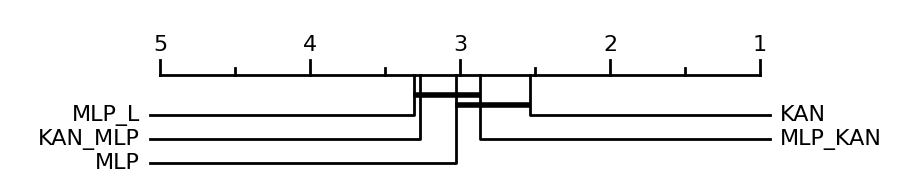} 
    \caption{Critical diagram of accuracy for five models across 128 datasets (higher rank is better)} 
    \label{fig:cd} 
\end{figure}

Fig. \ref{fig:Main_Result} (a) and (b) show the accuracy and F1 distribution across 128 UCR datasets for the five models respectively. We observe that these five models achieve relatively similar performance across the 128 datasets, both in terms of F1 score and accuracy. However, KAN performs slightly better overall. This conclusion is also supported by the results shown from the critical diagrams in Fig. \ref{fig:cd}, where only KAN and  MLP\_L in the same parameters exhibit statistically significant differences. In the critical diagram, KAN ranks the highest, indicating its strong fitting capability and demonstrating that it can achieve performance comparable to, or even better than, traditional neural networks on the benchmark time series datasets.

\subsection{Ablation Study of KAN}
KAN have a more complex structure compared to MLP, due to the combinations of base and spline functions. The different grid sizes of spline functions have varying impacts on performance. To evaluate their influences, we investigated three configurations of KAN as follows:

\begin{enumerate}
    \item Complete KAN with different grid sizes: 1, 5, 50
    \item KAN with only the wpline component, with different grid sizes: 1, 5, 50
    \item KAN with only the base component
\end{enumerate}
\begin{table}[h]
\centering
\caption{Test accuracy of models with different architectures on the 128 dataset. The values corresponding to columns 1, 5, and 50 represent the number of datasets out of 128 where the accuracy of the grid size corresponding to each row is greater than or equal to that of the grid size in the column, under the same architecture. Q1, Q2, and Q3 denote the quantiles of the accuracy distribution across the 128 datasets.}
\vspace{5pt} 
\label{tab:grid_study}
\setlength{\tabcolsep}{4pt} 
\renewcommand{\arraystretch}{1.2} 
\begin{tabular}{l|c|ccc|ccc}
\toprule
\textbf{KAN Configuration} & \textbf{Grid Size} & \textbf{1} & \textbf{5} & \textbf{50} & \textbf{Q1} & \textbf{Q2} & \textbf{Q3} \\ 
\midrule
\multirow{3}{*}{\textbf{w/ base \& Spline}} & 1 & 128 & 64 & 96 & 0.6000 & 0.7991 & 0.9214 \\ \cline{2-8} 
                                   & 5 & 76 & 128 & 112 & 0.6146 & 0.7750 & 0.9387 \\ \cline{2-8} 
                                   & 50 & 39 & 24 & 128 & 0.5626 & 0.6976 & 0.847 \\ \hline
\multirow{3}{*}{\textbf{w/o Base}}          & 1 & 128 & 100 & 122 & 0.5591 & 0.7571 & 0.9009 \\ \cline{2-8} 
                                   & 5 & 40 & 128 & 119 & 0.5054 & 0.6706 & 0.8271 \\ \cline{2-8} 
                                   & 50 & 13 & 19 & 128 & 0.2226 & 0.4315 & 0.5732\\ \hline
\textbf{w/o Spline }                        &-& \multicolumn{3}{c|}{-}  &0.5652& 0.7698 & 0.9000 \\ \hline
\end{tabular}
\end{table}

\begin{table}[h]
\centering
\caption{Test(Train) accuracy of different KAN  on the \textbf{CBF} dataset.}
\label{tab:CBF}
\setlength{\tabcolsep}{4pt} 
\renewcommand{\arraystretch}{1.2} 
\begin{tabular}{l|ccc}
\toprule
\textbf{KAN Configuration} & \textbf{1} & \textbf{5} & \textbf{50}  \\
\midrule
\textbf{w/ Base \& Spline}     & 0.9011(0.9667)    & 0.9644(0.9667) & 0.9300(0.9667)  \\
\textbf{w/o Base}   & 0.8722(1.0000)    & 0.8644(0.9667) & 0.3178(0.6667)  \\
\textbf{w/o Spline} & 0.8811(1.0000)  &    0.8811(1.0000)   &  0.8811(1.0000) \\
\bottomrule
\end{tabular}
\end{table}

\begin{figure}[h]
    \centering
    \begin{subfigure}[b]{0.475\textwidth}
        \centering
        \includegraphics[width=\textwidth]{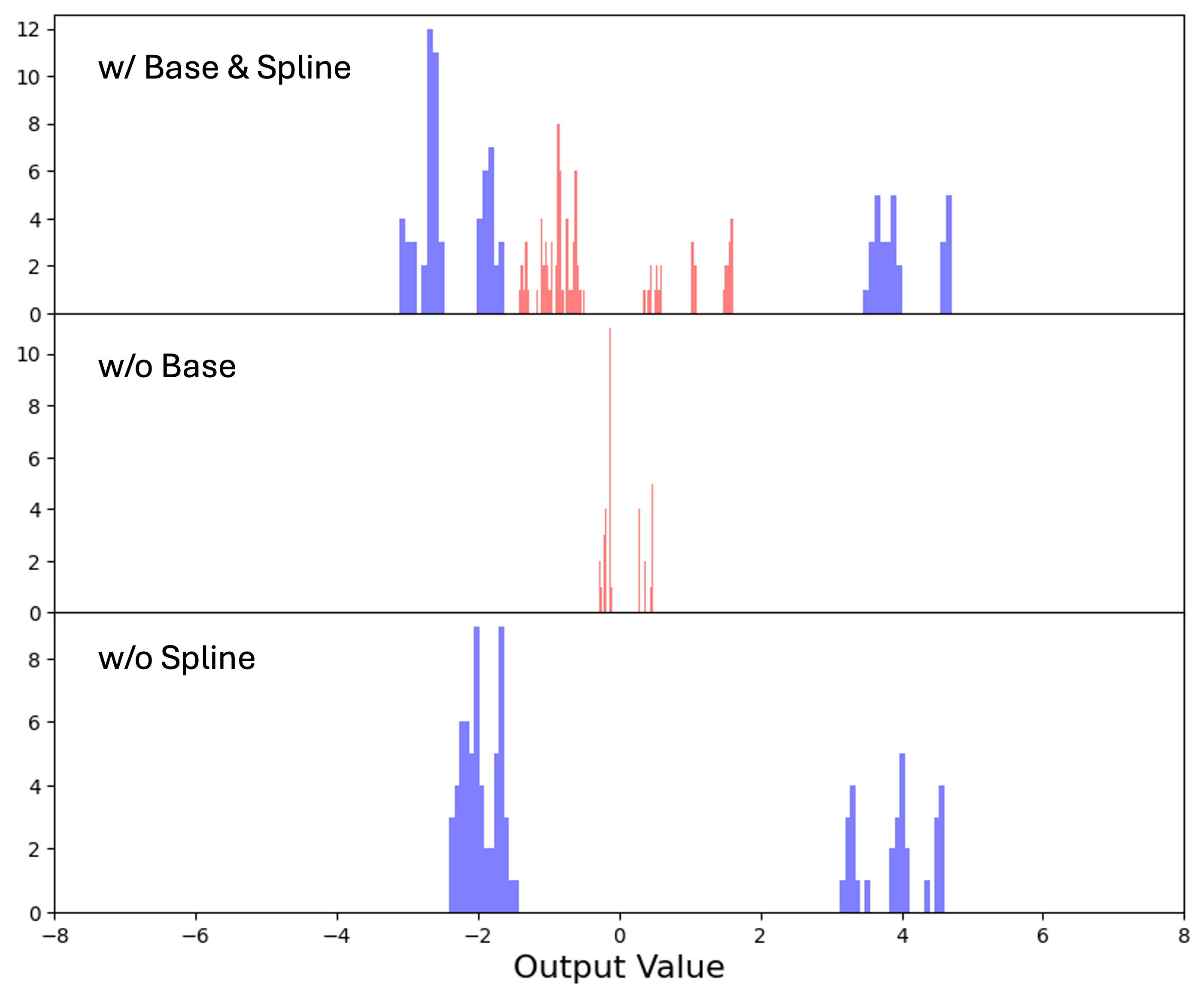} 
        \caption{Train: Grid Size = 1} 
        \label{fig:output_analysis1train} 
    \end{subfigure}
    \hspace{0pt}
    \begin{subfigure}[b]{0.475\textwidth}
        \centering
        \includegraphics[width=\textwidth]{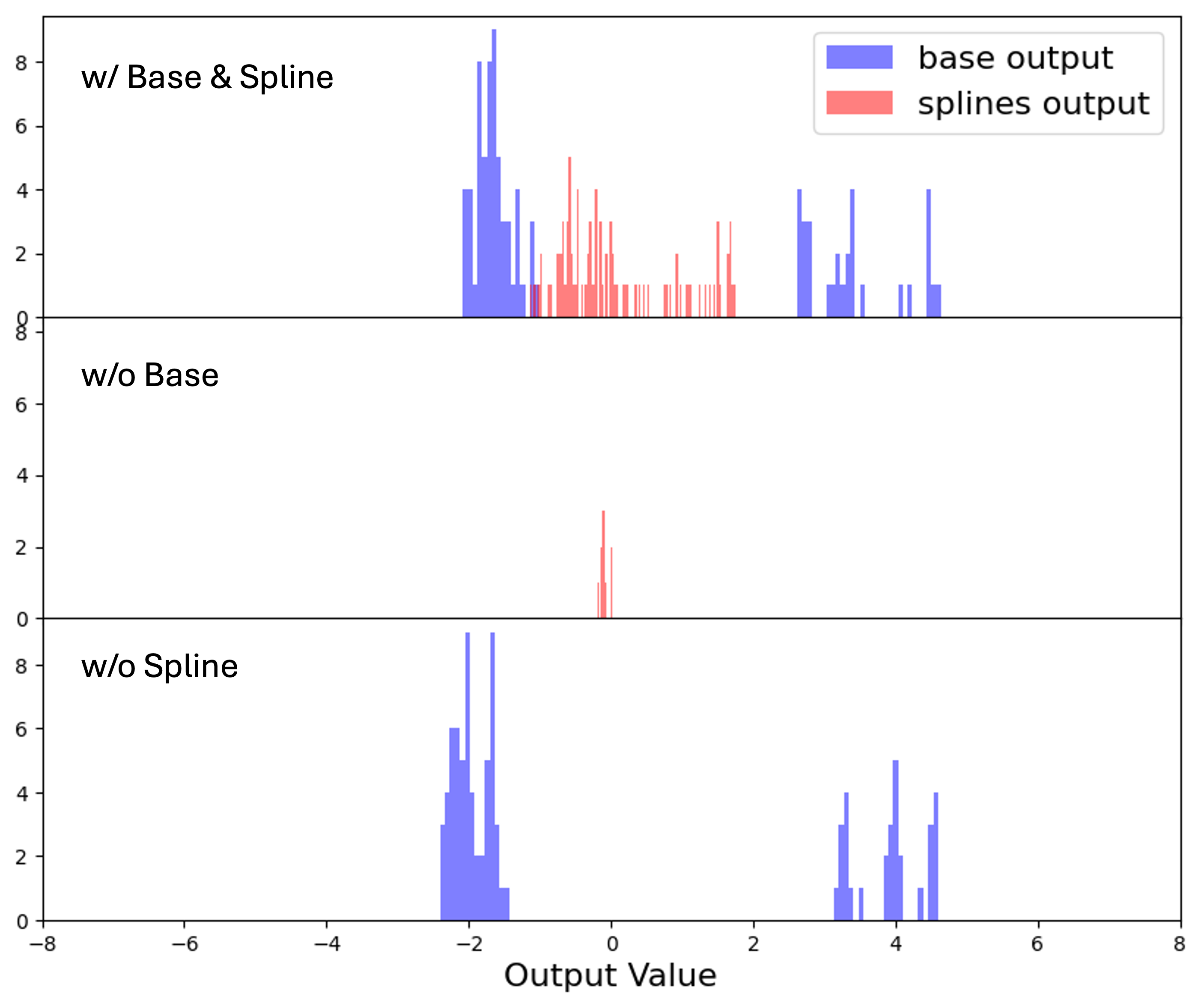} 
        \caption{Train: Grid Size = 50} 
        \label{fig:output_analysis2train} 
    \end{subfigure}
    \begin{subfigure}[b]{0.475\textwidth}
        \centering
        \includegraphics[width=\textwidth]{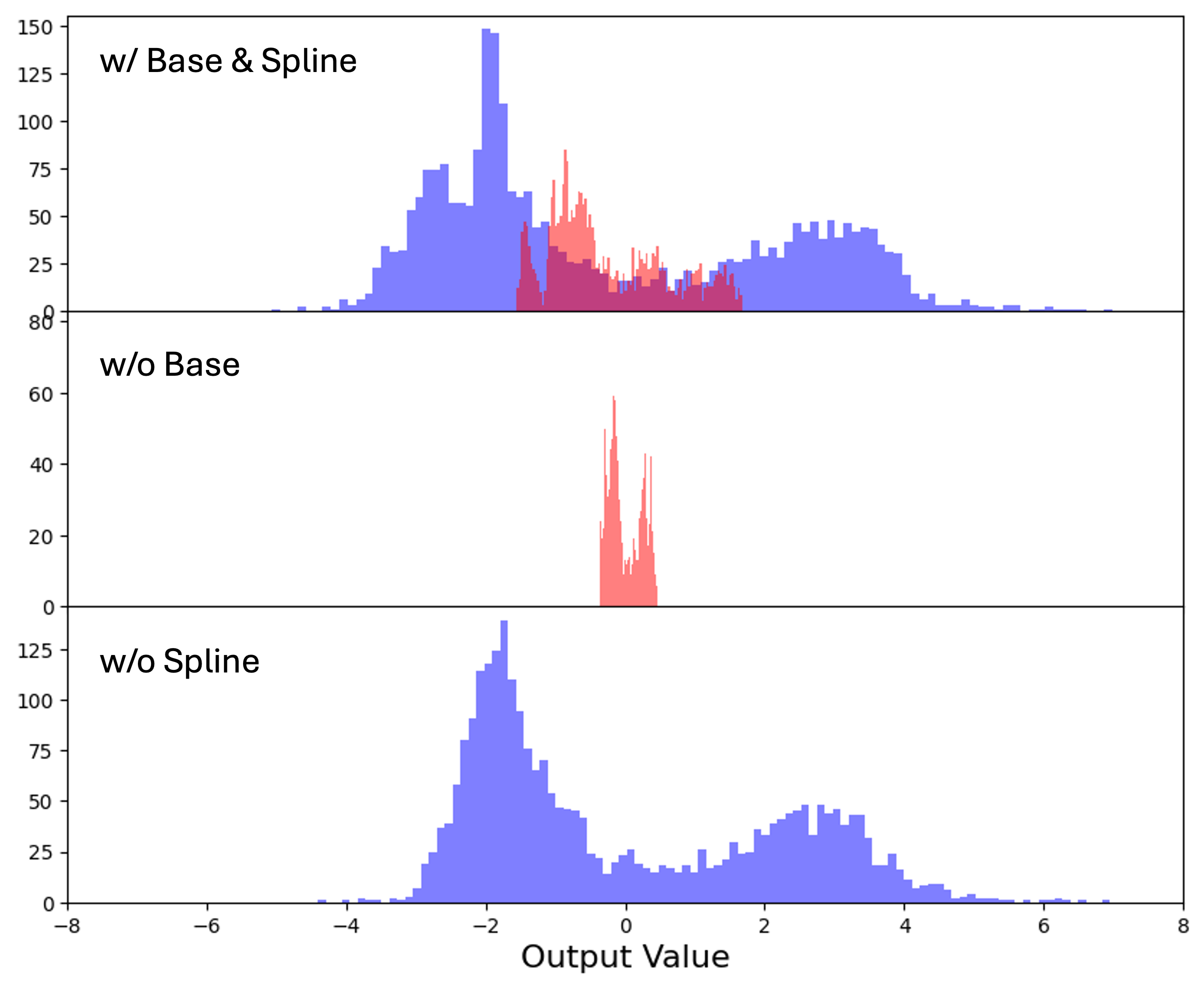}
        \caption{Test: Grid Size = 1}
        \label{fig:output_analysis1}
    \end{subfigure}
    \hspace{0pt}
    \begin{subfigure}[b]{0.475\textwidth}
        \centering
        \includegraphics[width=\textwidth]{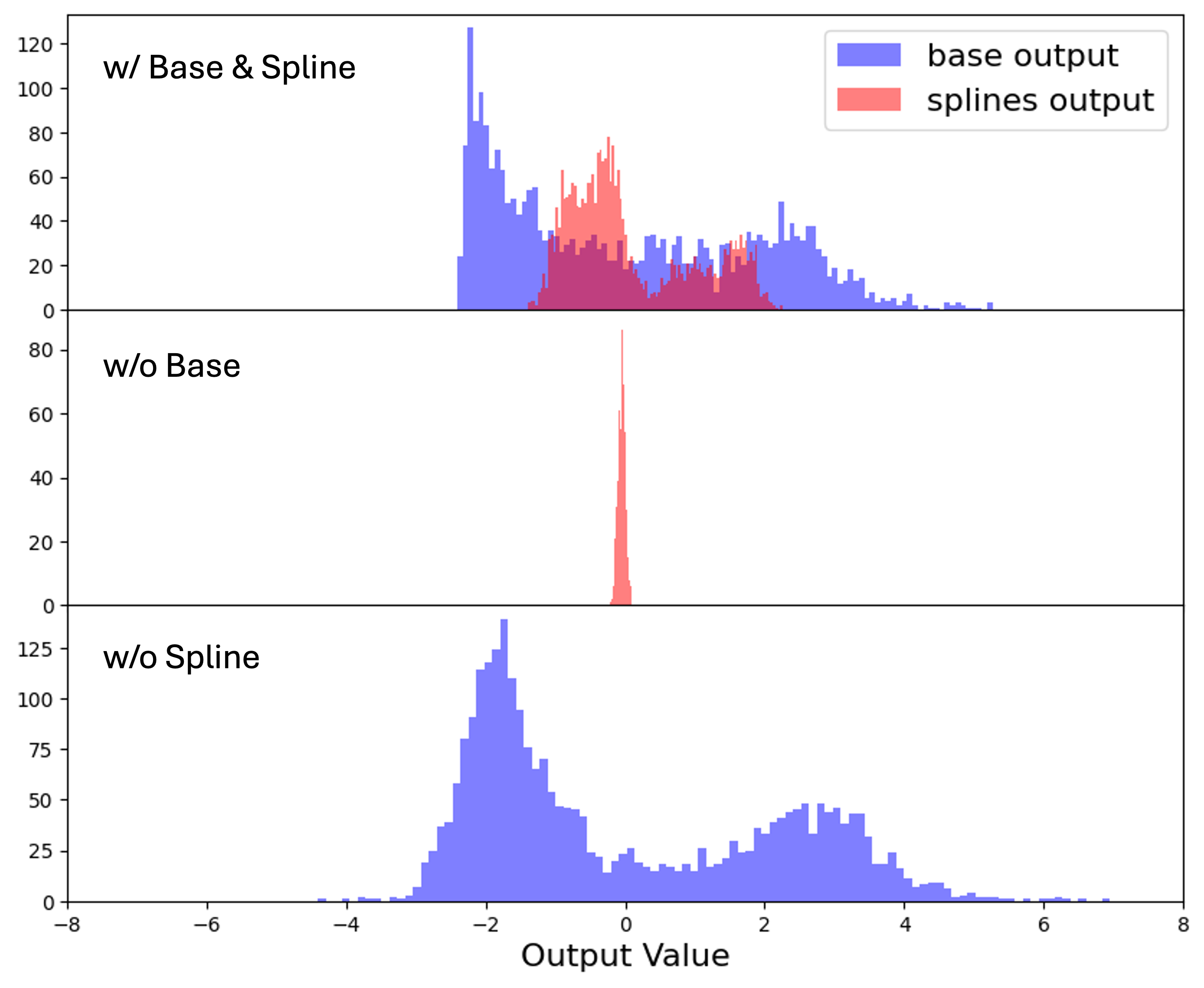}
        \caption{Test: Grid Size = 50}
        \label{fig:output_analysis2}
    \end{subfigure}

    \caption{Distribution of the flattened Train/Test output values of the last layer of the model under different configurations on the \textbf{CBF} dataset. (a) Train: Grid size of 1, (b)Train:  Grid size of 50, (c) Test: Grid size of 1, and (d)Test: Grid size of 50.}
    \label{fig:output_analysis}
\end{figure}

\begin{figure}[h]
    \centering
    
    \begin{subfigure}[b]{0.475\textwidth}
        \centering
        \includegraphics[width=\textwidth]{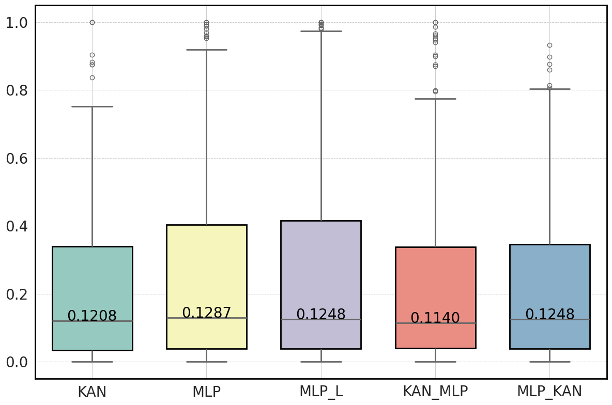} 
        \caption{eps = 0.05} 
        \label{fig:ASR_eps_0.05} 
    \end{subfigure}
    \hspace{0pt}
    \begin{subfigure}[b]{0.475\textwidth}
        \centering
        \includegraphics[width=\textwidth]{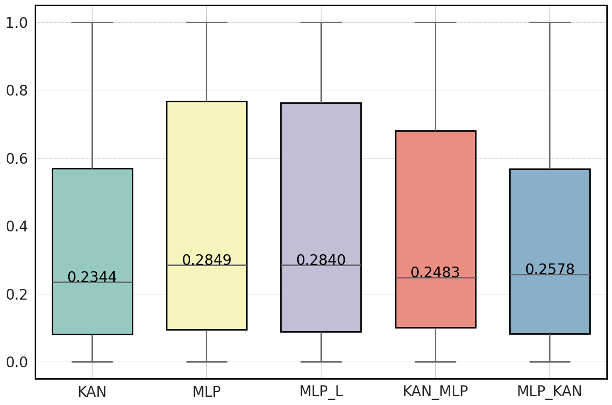} 
        \caption{eps = 0.1} 
        \label{fig:ASR_eps_0.1} 
    \end{subfigure}
    \begin{subfigure}[b]{0.475\textwidth}
        \centering
        \includegraphics[width=\textwidth]{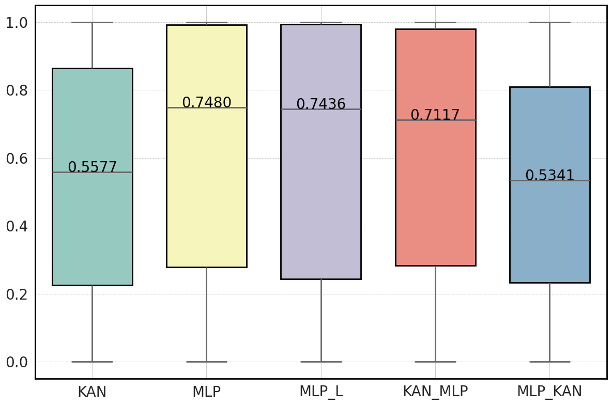}
        \caption{eps = 0.25}
        \label{fig:ASR_eps_0.25}
    \end{subfigure}
    \hspace{0pt}
    \begin{subfigure}[b]{0.475\textwidth}
        \centering
        \includegraphics[width=\textwidth]{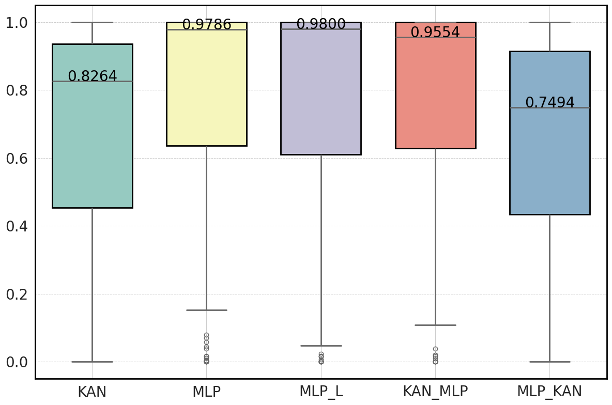}
        \caption{eps = 0.5}
        \label{fig:ASR_eps_0.5}
    \end{subfigure}

    \caption{ASR distribution across 128 datasets for five models in different perturbation eps.}
    \label{fig:ASR}
\end{figure}

\begin{figure}[h]
    \centering
    
    \begin{subfigure}[b]{0.475\textwidth}
        \centering
        \includegraphics[width=\textwidth]{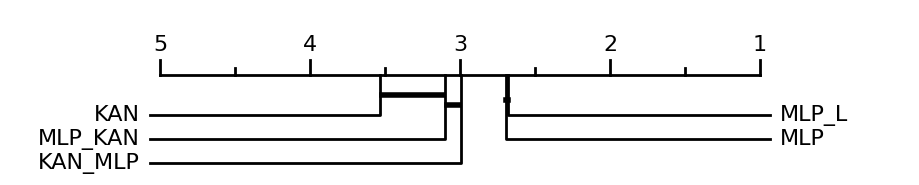} 
        \caption{eps = 0.05} 
        \label{fig:ASR_eps_0.05} 
    \end{subfigure}
    \hspace{0pt}
    \begin{subfigure}[b]{0.475\textwidth}
        \centering
        \includegraphics[width=\textwidth]{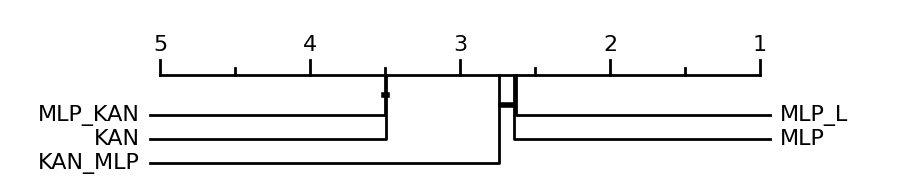} 
        \caption{eps = 0.1} 
        \label{fig:ASR_eps_0.1} 
    \end{subfigure}
    \begin{subfigure}[b]{0.475\textwidth}
        \centering
        \includegraphics[width=\textwidth]{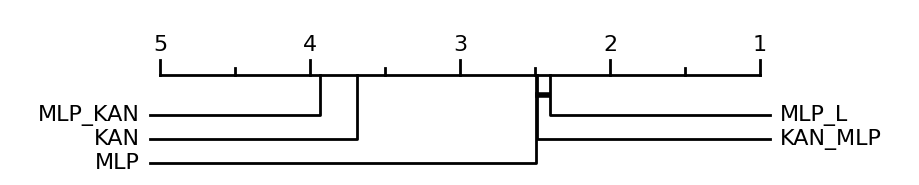}
        \caption{eps = 0.25}
        \label{fig:ASR_eps_0.25}
    \end{subfigure}
    \hspace{0pt}
    \begin{subfigure}[b]{0.475\textwidth}
        \centering
        \includegraphics[width=\textwidth]{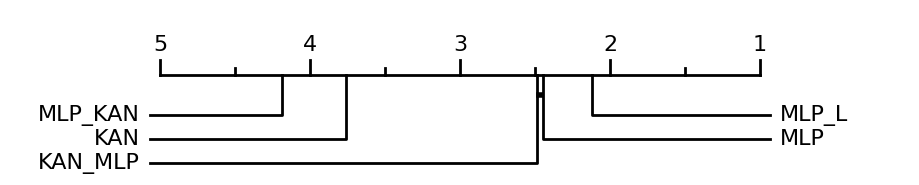}
        \caption{eps = 0.5}
        \label{fig:ASR_eps_0.5}
    \end{subfigure}

    \caption{Critical diagram of ASR rank across 128 datasets for five models in different perturbation eps (lower
rank is better)}
    \label{fig:ASR_CD}
\end{figure}

\begin{figure}[h]
    \centering
    \begin{subfigure}[b]{0.45\textwidth}
        \centering
        \includegraphics[width=\textwidth]{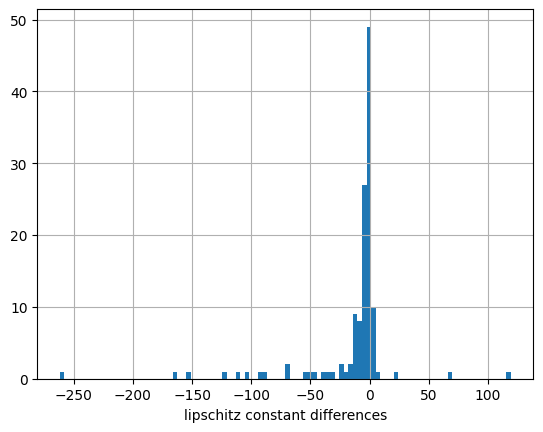} 
        \caption{KAN - MLP} 
        \label{fig:lip_diff_kan_ml} 
    \end{subfigure}
    \hfill
    \begin{subfigure}[b]{0.45\textwidth}
        \centering
        \includegraphics[width=\textwidth]{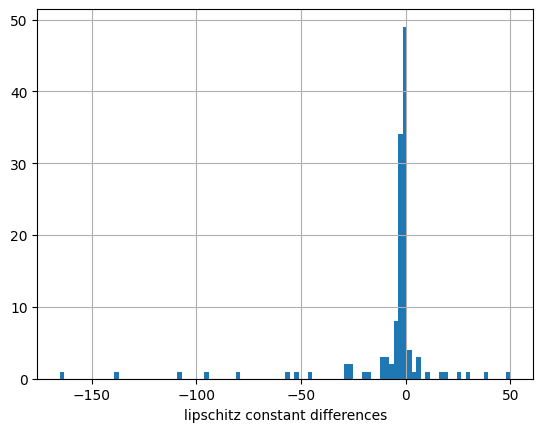} 
        \caption{MLP\_KAN - MLP} 
        \label{fig:lip_diff_mlp_kan_mlp} 
    \end{subfigure}
    \caption{Lipschitz constant distribution differences across 128 dataset} 
    \label{fig:Lip_diff_ALL} 
\end{figure}

\begin{figure}[!h]
    \centering
    \begin{subfigure}[b]{0.475\textwidth}
        \centering
        \includegraphics[width=\textwidth]{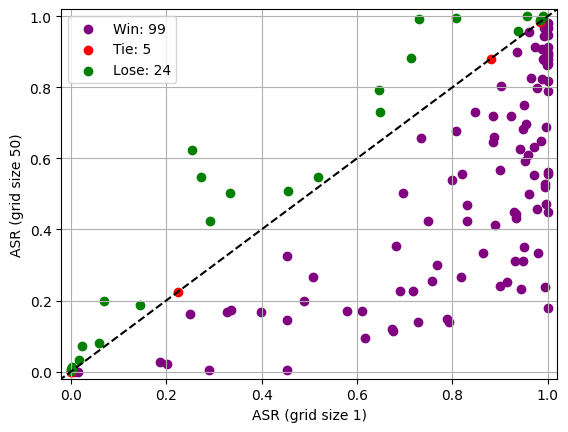} 
        \caption{Grid Size 50 vs Grid Size 1} 
        \label{fig:lip_diff_kan_ml} 
    \end{subfigure}
    \hfill
    \begin{subfigure}[b]{0.455\textwidth}
        \centering
        \includegraphics[width=\textwidth]{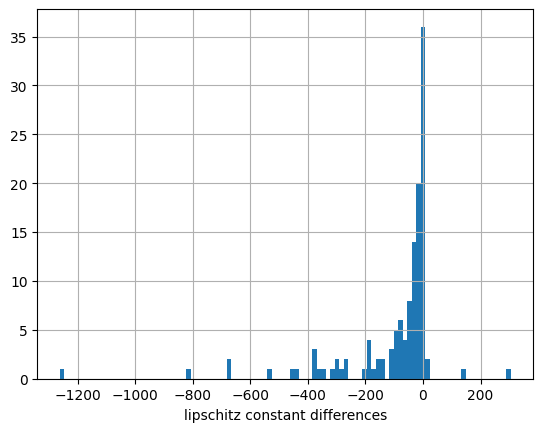} 
        \caption{Grid Size 1 - Grid Size 50} 
        \label{fig:lip_diff_mlp_kan_mlp} 
    \end{subfigure}
    \caption{Comparison of (a) ASR and (b) Lipschitz constant distribution differences across 128 dataset between Grid Size of 50 and Grid Size of 1} 
    \label{fig:gs50_gs1} 
\end{figure}
Tab.~\ref{tab:grid_study} presents the overall performance of these three configurations across 128 datasets. We observed that an excessively large grid size leads to performance degradation, regardless of whether it is in the complete KAN or without the base function. In the complete KAN, there is little difference in performance with smaller grid sizes. For grid size = 1, nearly 50\% of the datasets achieved over 80\% accuracy, whereas for grid size = 50, this value drops to less than 70\%. In the KAN without the base function, overall performance significantly declines as the grid size increases. Particularly, for grid size = 50, the accuracy of 50\% of the datasets is below 43.15\%. Additionally, we found that the performance of KAN without the spline function is close to that of the KAN network with only the spline function and with grid size = 1. This indicates that the fitting capability of KAN largely comes from the simple activation functions, suggesting that complex B-spline combinations may lead to optimization difficulties.
To explain the result above, we analyzed the \textbf{CBF} dataset, which exhibited results similar to the overall trend as shown in tab.~\ref{tab:CBF}. Most results are comparable, except for the model with only the Spline function and a grid size of 50, which performed significantly worse. We analyzed the output results of the two parts at the last layer as shown in the fig.~\ref{fig:output_analysis}.

We observed two phenomena across both the training and testing sets: First, the output values of the spline are relatively smaller and more concentrated compared to those of the base configuration. This indicates that the spline's contribution to the final decision is less significant than that of the base, suggesting that the base configuration plays a more critical role in decision-making. Second, the most significant difference between fig.~\ref{fig:output_analysis1} and \ref{fig:output_analysis2} is the distinct distribution observed when the base component is removed. When the grid size is set to 1, the output distributions for these three configurations are similar, exhibiting two prominent peaks on both the positive and negative sides, with the negative peak being higher and more numerous. This pattern occurs because the \textbf{CBF} dataset has 3 categories, thus, when one class is predicted with high confidence, the other two classes tend to output negative values. However, this scenario changes drastically at a grid size of 50. Here, the spline output shows only a single peak concentrated around zero both in the training and testing set, corresponding to the lower accuracy observed in tab.~\ref{tab:grid_study} for a grid size of 50 (without Base). This further confirms that an excessively large grid size complicates the network's optimization.
\subsection{Evaluation and Analysis of Adversarial Robustness}

We also found that KAN demonstrate better robustness compared to MLP. We performed PGD untargeted attacks on the aforementioned five models, with \(\epsilon\) ranging from 0.05 to 0.5. The results consistently show that KAN significantly outperform MLP. Fig.~\ref{fig:ASR} illustrates the ASR of PGD on these models. We observe that KAN and MLP\_KAN exhibit remarkable robustness compared to the other three models, with this advantage increasing as \(\epsilon\) grows. Specifically, at \(\epsilon = 0.5\), the MLP\_KAN model shows the best robustness among the five, with the ASR on 50\% of the dataset remaining below approximately 75\%, whereas the ASR for MLP, MLP\_L, and KAN\_MLP approach 1 for nearly 50\% of the dataset. From fig.~\ref{fig:ASR_CD}, it is evident that KAN and MLP\_KAN demonstrate significantly different robustness across 128 datasets compared to the other three models.

To explain this phenomenon, we obtained the Lipschitz constants for the KAN, MLP, and MLP\_KAN models. Fig.~\ref{fig:Lip_diff_ALL} shows the distribution of Lipschitz constant differences across 128 datasets. Fig.~\ref{fig:lip_diff_kan_ml} and \ref{fig:lip_diff_mlp_kan_mlp} both indicate that the model with KAN layer generally results in a decrease in the Lipschitz constants for most datasets, which is consistent with the observations in fig.~\ref{fig:ASR}. The combination of spline functions produced by KAN with a low grid size tends to be smooth and flat, making it difficult for small changes in input to cause significant changes in the output \(y\). Additionally, as shown in the previous fig.~\ref{fig:output_analysis}, the output distribution of the B-spline function is more narrow compared to the combination of activation and linear functions, thus contributing minimally to the output. This could be the primary reason why KAN is more robust under attack.

However, we observed the opposite result in another experiment. Fig.~\ref{fig:gs50_gs1} shows that the Lipschitz constant corresponding to a larger grid size is greater. This is evident because as the number of grids increases, the slopes of the spline basis also increase, making the overall B-spline function more likely to produce larger slopes. However, the results indicate that a larger Lipschitz constant leads to greater robustness, with 104 out of 128 datasets having an ASR less than or equal to that corresponding to a grid size of 1, demonstrating stronger robustness. We preliminarily believe this might be because the value distribution produced by the B-spline is much smaller compared to the base function, thus the base contributes the majority in decision-making. However, the gradients of the B-spline with a larger grid size are substantial, thus during PGD attacks, the gradient is mainly provided by the B-spline part, which might not necessarily provide the correct direction compared to the base. Therefore, networks with larger Lipschitz constants exhibit stronger robustness. This may also further imply why models are difficult to optimize as the grid size increases.

\newpage
\section{Conclusion}
In this paper, we applied the KAN in Time Series Classification and conducted a fair comparison among KAN, MLP, and mixed structures. We found that KAN can achieve comparable performance to MLP. Additionally, we analyzed the importance of each component of KAN and discovered that a larger grid size can be difficult to optimize, leading to lower performance. Furthermore, we evaluated the adversarial robustness of KAN and these models, finding that KAN exhibited remarkable robustness. This robustness is attributed to KAN's lower Lipschitz constant. Moreover, we found that KAN with a larger grid size have a greater Lipschitz constant but exhibit stronger robustness. We provided an explanation for this phenomenon, although it requires further verification and broader experiments in our future work.

\section{Acknowledgments}
We thank all the creators and providers of the UCR time series
benchmark datasets. This research work is supported by Australian Research Council Linkage Project (LP230200821), Australian
Research Council Discovery Projects (DP240103070 ), Australian Research Council ARC Early Career Industry Fellowship (IE230100119),
Australian Research Council ARC Early Career Industry Fellowship
(IE240100275), and University of Adelaide, Sustainability FAME Strategy Internal Grant 2023.

\newpage
\bibliographystyle{splncs04}
\bibliography{mybibliography}

\begin{thebibliography}{10}
\providecommand{\url}[1]{\texttt{#1}}
\providecommand{\urlprefix}{URL }
\providecommand{\doi}[1]{https://doi.org/#1}

\bibitem{bagnall2017great}
Bagnall, A., Lines, J., Bostrom, A., Large, J., Keogh, E.: The great time
  series classification bake off: a review and experimental evaluation of
  recent algorithmic advances. Data mining and knowledge discovery
  \textbf{31},  606--660 (2017)

\bibitem{dau2019ucr}
Dau, H.A., Bagnall, A., Kamgar, K., Yeh, C.C.M., Zhu, Y., Gharghabi, S.,
  Ratanamahatana, C.A., Keogh, E.: The ucr time series archive. IEEE/CAA
  Journal of Automatica Sinica  \textbf{6}(6),  1293--1305 (2019)

\bibitem{dong2024boosting}
Dong, C., Li, Z., Zheng, L., Chen, W., Zhang, W.E.: Boosting certificate
  robustness for time series classification with efficient self-ensemble. arXiv
  preprint arXiv:2409.02802  (2024)

\bibitem{dong2023swap}
Dong, C.G., Zheng, L.N., Chen, W., Zhang, W.E., Yue, L.: Swap: Exploiting
  second-ranked logits for adversarial attacks on time series. In: 2023 IEEE
  International Conference on Knowledge Graph (ICKG). pp. 117--125. IEEE (2023)

\bibitem{fawaz2019adversarial}
Fawaz, H.I., Forestier, G., Weber, J., Idoumghar, L., Muller, P.A.: Adversarial
  attacks on deep neural networks for time series classification. In: 2019
  International Joint Conference on Neural Networks (IJCNN). pp.~1--8. IEEE
  (2019)

\bibitem{forestier2018surgical}
Forestier, G., Petitjean, F., Senin, P., Despinoy, F., Huaulm{\'e}, A., Fawaz,
  H.I., Weber, J., Idoumghar, L., Muller, P.A., Jannin, P.: Surgical motion
  analysis using discriminative interpretable patterns. Artificial intelligence
  in medicine  \textbf{91},  3--11 (2018)

\bibitem{hills2014classification}
Hills, J., Lines, J., Baranauskas, E., Mapp, J., Bagnall, A.: Classification of
  time series by shapelet transformation. Data mining and knowledge discovery
  \textbf{28},  851--881 (2014)

\bibitem{hornik1989multilayer}
Hornik, K., Stinchcombe, M., White, H.: Multilayer feedforward networks are
  universal approximators. Neural networks  \textbf{2}(5),  359--366 (1989)

\bibitem{ismail2020inceptiontime}
Ismail~Fawaz, H., Lucas, B., Forestier, G., Pelletier, C., Schmidt, D.F.,
  Weber, J., Webb, G.I., Idoumghar, L., Muller, P.A., Petitjean, F.:
  Inceptiontime: Finding alexnet for time series classification. Data Mining
  and Knowledge Discovery  \textbf{34}(6),  1936--1962 (2020)

\bibitem{karim2020adversarial}
Karim, F., Majumdar, S., Darabi, H.: Adversarial attacks on time series. IEEE
  transactions on pattern analysis and machine intelligence  \textbf{43}(10),
  3309--3320 (2020)

\bibitem{karim2017lstm}
Karim, F., Majumdar, S., Darabi, H., Chen, S.: Lstm fully convolutional
  networks for time series classification. IEEE access  \textbf{6},  1662--1669
  (2017)

\bibitem{kolmogorov1961representation}
Kolmogorov, A.N.: On the representation of continuous functions of several
  variables by superpositions of continuous functions of a smaller number of
  variables. American Mathematical Society  (1961)

\bibitem{7837946}
Lines, J., Taylor, S., Bagnall, A.: Hive-cote: The hierarchical vote collective
  of transformation-based ensembles for time series classification. In: 2016
  IEEE 16th International Conference on Data Mining (ICDM). pp. 1041--1046
  (2016). \doi{10.1109/ICDM.2016.0133}

\bibitem{liu2024kan}
Liu, Z., Wang, Y., Vaidya, S., Ruehle, F., Halverson, J., Solja{\v{c}}i{\'c},
  M., Hou, T.Y., Tegmark, M.: Kan: Kolmogorov-arnold networks. arXiv preprint
  arXiv:2404.19756  (2024)

\bibitem{madry2017towards}
Madry, A., Makelov, A., Schmidt, L., Tsipras, D., Vladu, A.: Towards deep
  learning models resistant to adversarial attacks. arXiv preprint
  arXiv:1706.06083  (2017)

\bibitem{nweke2018deep}
Nweke, H.F., Teh, Y.W., Al-Garadi, M.A., Alo, U.R.: Deep learning algorithms
  for human activity recognition using mobile and wearable sensor networks:
  State of the art and research challenges. Expert Systems with Applications
  \textbf{105},  233--261 (2018)

\bibitem{pelletier2019temporal}
Pelletier, C., Webb, G.I., Petitjean, F.: Temporal convolutional neural network
  for the classification of satellite image time series. Remote Sensing
  \textbf{11}(5), ~523 (2019)

\bibitem{schafer2015boss}
Sch{\"a}fer, P.: The boss is concerned with time series classification in the
  presence of noise. Data Mining and Knowledge Discovery  \textbf{29},
  1505--1530 (2015)

\bibitem{tholeti2022robust}
Tholeti, T., Kalyani, S.: The robust way to stack and bag: the local lipschitz
  way. arXiv preprint arXiv:2206.00513  (2022)

\bibitem{wen2022transformers}
Wen, Q., Zhou, T., Zhang, C., Chen, W., Ma, Z., Yan, J., Sun, L.: Transformers
  in time series: A survey. arXiv preprint arXiv:2202.07125  (2022)

\bibitem{zhou2023one}
Zhou, T., Niu, P., Sun, L., Jin, R., et~al.: One fits all: Power general time
  series analysis by pretrained lm. Advances in neural information processing
  systems  \textbf{36},  43322--43355 (2023)

\end{thebibliography}

\end{document}